# Satisfaction of Assumptions is a Weak Predictor of Performance

Ben P. Wise*
Thayer School, Dartmouth
Hanover, NH 03755

## 1. Abstract

This paper demonstrates a methodology for examining the accuracy of uncertain inference systems (UIS), after their parameters have been optimized, and does so for several common UIS's. This methodology may be used to test the accuracy when either the prior assumptions or updating formulae are not exactly satisfied. Surprisingly, these UIS's were revealed to be no more accurate on the average than a simple linear regression. Moreover, even on prior distributions which were deliberately biased so as give very good accuracy, they were less accurate than the simple probabilistic model which assumes *marginal* independence between inputs. This demonstrates that the importance of updating formulae can outweigh that of prior assumptions. Thus, when UIS's are judged by their final accuracy after optimization, we get completely different results than when they are judged by whether or not their prior assumptions are perfectly satisfied.

## 2. Introduction

This paper demonstrates a methodology for examining the behavior of uncertain inference systems (UIS), after their parameters have been optimized, and does so for several UIS's. Our criterion for "performance" is simply the accuracy of their answers. Of course, other criteria (e.g. speed of execution, modularity, ease of explanation, etc.) are very important in a real system. However, accuracy is still of fundamental importance: while we may be forced to use approximations in order to achieve speed, explicability, modularity, etc., we must also verify that the "approximation" actually is reasonably close to the ideal answer, after we have done everything we can to optimize its performance. Moreover, we would like to know if accuracy is maintained as the problem changes, i.e. whether the UIS is robust or brittle. In essence, we can not make careful trade-offs between speed, ease, and accuracy unless we have some way measuring each. This paper demonstrates a methodolgy for, and some results from, measuring accuracy. This orientation to final performance is in contrast to the usual approach, which analyzes the prior assumptions which went into their design, and ranks UIS's by how plausible or flexible those assumptions are. Our approach is motivated by the attitude that "I don't care what theory it uses; I just care whether it works or not". We will demonstrate that the process of optimization, and the shift of viewpoint from prior assumptions to final performance, completely changes the ranking. The analysis was done on six different systems, one of which is derived from PROSPECTOR. Although PROSPECTOR is no longer a new, cutting-edge system, and hence criticisms of it have the flavor of "beating a dead horse", it is an example which is both relatively sophisticated and relatively well-known. The fact that careful analysis of PROSPECTOR gives surprising and counter-intuitive results indicates that the *newer* systems may also turn out to violate intuition. Moreover, a widely-rejected but exact model was discovered to be quite accurate and robust. Thus, we have two practical implications. First, careful, explicit analysis of heuristic UIS's performance after optimization is highly recommended, because performance is simply not well-predicted by prior assumptions. Second, even when a UIS is based on an exact probabilistic model, we need to assess its robustness when that model's assumptions are violated.

## 3. Definition of UIS's Examined

We selected a small set of UIS's to study, and assessed their behavior over a large set of problems. The UIS's considered were EMYCIN, PROSPECTOR ($PRSP$), a simple in-

---

dependence model ($INDP$), linear regression of the logarithm of the odds-ratios ($PWR$), and linear regression of probabilities ($LINR$). The situation considered was one in which evidence from two separate rules bore on one conclusion. Both EMYCIN and PROSPECTOR provide "Bayesian" combination rules for this situation, as well as the usual "and" and "implication" rules for several pieces of evidence which go through one rule to bear on a conclusion. The latter situation can be algebraically analyzed, as in [6]. The "Bayesian" rules are too complex for easy algebraic analysis, and so our numerical explorations concentrated on the "Bayesian" rules. Given this restricted set of rules, the only thing left to vary was the prior probability distribution (the "underlying distribution") which the rule was supposed to model. We will first define each model, then describe the experimental design and optimization procedure.

The $LINR$ model predicts the posterior probability of $C$ as a linear function of the new posterior probabilities for $E1$ and $E2$. In the $i$-th data set, the new probability of $Ej$ is denoted by $e_{ji}$.

$$c_{LINR} = a_1 e_{1i} + a_2 e_{2i} + b$$

The $INDP$ model is defined simply by assuming that $E1$ and $E2$ are *marginally* independent, and calculating the exact odds-ratio update under that assumption. The important point to note here is that while we make an approximation in assuming the independent prior, the update is exact. Hence, for the $INDP$ model, there is no error due to updating, and errors can arise only from the approximate prior assumptions. Algebraically, one finds that there are four missing parameters, which correspond to the various conditional probabilities of $C$ given different combinations of $E1$ and $E2$. This gives equation (3),

$$c_{INDP} = b_{00}(1 - e_{1i})(1 - e_{2i}) + b_{01}(1 - e_{1i})(e_{2i}) + b_{10}(e_{1i})(1 - e_{2i}) + b_{11}(e_{1i})(e_{2i}) \quad (3)$$

In probabilistic problems, the $INDP$ model can be seen as analogous to the linear regression model in statistical problems. It is a simple model which can be exactly solved. All the issues relating to its use deal with how robust it is when the underlying assumptions are violated. In statistical problems, we know that real-world relationships are almost never exactly linear. Nevertheless, it is a widely observed fact that the linear model *after optimization* is frequently a very good model, often out-performing human experts. At least by analogy, this raises the possibility that the independence model may, after optimization, be quite accurate.

The $PRSP$ model is just the probabilistic updating equation from PROSPECTOR, which assumes *conditional* independence of $E1$ and $E2$ given $C$, and also given $\neg C$. It gives exactly the correct answer when $e_{1i}$ and $e_{2i}$ are both either 0 or 1. A heuristic interpolation procedure is used when $e_{1i}$ or $e_{2i}$ lie between 0 and 1 [2]. If one finds a closed-form answer for the output of the procedure, one discovers that the probabilities of $E1$ and $E2$ from the prior distribution appear as constants in the final answer. In our optimization, we treated each appearance of these probabilities as independent parameters to the formula. Hence, our updating formula has more degrees of freedom, and greater flexibility, than the procedure actually used in PROSPECTOR. As always in optimizations, this extra flexibility will improve the optimized performance.

The third model, $PWR$, is a linear regression of the logarithms of the odds-ratios, as in equation (5).

$$\ln\left(\frac{c_{PWR}}{1 - c_{PWR}}\right) = a_1 \ln\left(\frac{e_{1i}}{1 - e_{1i}}\right) + a_2 \ln\left(\frac{e_{2i}}{1 - e_{2i}}\right) + b \quad (5)$$

One might expect the $PWR$ model to do rather well; in fact, it was consistently less accurate than the simple linear model. The reasons for this odd result remain unknown.

## 4. Experimental Method

Assessing the final accuracy of optimized UIS's is an analytically intractable problem. Hence, we approached it by numerical optimization, repeated for many different small inference problems. Each inference problem was represented by a distribution over three events: two pieces of evidence, $E1$ and $E2$, and a conclusion, $C$. As each can be either *true* of *false*, there were eight entries in each distribution. Distributions were generated in two sets. The first set was formed by sampling over the whole space of possible distributions; the second was biased so as to favor $PRSP$ to the greatest possible degree. For the first set of distributions, we assured a broad and unbiased selection of rule-sets by generating the underlying distributions by uniform selection over the seven-dimensional space[1] of possible distributions.[2] The random sampling procedure assured that a broad range of problem-features would be represented The motivation here was to sample of broad range of problems so that we could later separate out the sets on which each UIS performed particularly well or poorly. For each rule-set generated, we individually optimized each UIS for optimal average performance on that particular rule-set. The results of this preliminary scan were surprising enough that we performed another scan which was deliberately biased so as to favor $PRSP$. This second set was generated by picking parameters (of which there are five) for distributions which exactly satisfy PROSPECTOR's conditional independence assumptions, and expanding out the corresponding distribution.

Our criterion both for optimizing UIS's and for ranking their performance was the root of the mean squared error, as compared to a certain standard. The standard we chose is the minimum cross-entropy, or odds-ratio, update. This standard was chosen for several reasons, as discussed in [7]. First, any coordinate-invariant method of updating will give the same answer as odds-ratio updating. Second, many popular updating schemes (ordinary Bayesian conditioning, Jeffrey's Rule, Bayes' Rule, Pearl's Bayesian Networks, and others) are each identical to odds-ratio updating when it is applied to their special cases.[3]

In the case of certain evidence, the minimum cross-entropy techniques used to generate the "standard answer" are identical to ordinary Bayesian conditioning, and hence can be used on virtually any distribution. This is useful for analyzing the robustness of inference schemes which update exactly, given priors with the appropriate characteristics. We will do this for the $INDP$ model; it can just as easily be done for the Bayes' Network formalism of Pearl [4]. This is entirely analogous to analysis of the classical linear regression model: it is exact when the assumptions are met, and the large concerns center about robustness when they are violated.

For each distribution, and each UIS, the RMS error was assessed by scanning the probabilities of $E1$ and $E2$ over (.001, .25, .5, .75, .999) independently, giving 25 combinations of $(e_{1i}, e_{2i})$. For each such pair, the "true" or standard value of $c_i$ was calculated by odds-ratio updating, giving a list of 25 numbers. Let us denote the list of input parameters for each UIS, $X$, by $P_X$. The estimate for $c$, on the $i$-th pair would thus be denoted by $\hat{c}_X(P_X, i)$. The overall RMS error for $X$, using the parameters $P_X$, is $\epsilon_X(P_X)$, as in equation (1)

$$\epsilon_X^2(P_X) = \sum_i (c_i - \hat{c}_X(P_X, i))^2 \qquad (1)$$

The optimization was performed with a deflected-gradients search, varying $P_X$ so as to minimize $\epsilon_X(P_X)$; this minimum value is denoted by $\epsilon_X$.[4] To arrive at some overall performance measure for a UIS, the 109 different $\epsilon$'s, one for each distribution, must somehow be combined. However, different probability distributions present differing degrees of complexity. In a trivial distribution, we might have two pieces of evidence, $E1$ and $E2$, which were independent of each other and of the conclusion, $C$. In this case, virtually any UIS would work perfectly. It is also possible for $E1$ and $E2$ to be *individually* independent of $C$, but for $E1$ and $E2$ in combination to be quite informative about $C$. On this latter case, most simple UIS's perform quite poorly. Thus, a low $\epsilon_X$ could be due to either to a very good UIS, or to a particularly easy problem. Hence, the $\epsilon_X$ value is then re-scaled to a performance measure by comparing it to the $\epsilon$-values for other UIS's on the same distribution.

---

[1] The space is seven-dimensional because there are eight probabilities, constrained to add up to 1.

[2] For a description of the non-trivial procedure for doing such a uniform sampling, see [7].

[3] See [7] for proofs.

To make the re-scaling intuitively meaningful, it was done relative to the best possible UIS, the worst possible optimized UIS, and a "straw man", the ordinary linear regression. The model $BST$ was defined to be the theoretically best possible, where $\hat{c}_i = c_i$. The model $WRST$ was to be the worst possible optimized model, i.e. one which ignored the evidence and hence guessed a constant. To minimize error, this constant was chosen to be simply the mean of the $c_i$. The model $LINR$ was defined earlier. To compare the performance of an arbitrary model to these three, we define the $\eta(X)$ measure in equation (2). Of course, a particular $\eta(X)$ value must be used when $X$ has been optimized for a specific distribution $D$, and the optimized versions of the others are used on the same distribution.

$$\eta(X) = \begin{cases} (\epsilon_X - \epsilon_{LINR}) \div (\epsilon_{BST} - \epsilon_{LINR}) & \text{if } \epsilon_X < \epsilon_{LINR} \\ (\epsilon_X - \epsilon_{LINR}) \div (\epsilon_{LINR} - \epsilon_{WRST}) & \text{if } \epsilon_X > \epsilon_{LINR} \end{cases} \quad (2)$$

Thus, any model which is more accurate than a simple linear regression will be given a positive $\eta$ score, while any model which is worse than simple linear regression will receive a negative score. For optimized models, scores will always range between +1, for the best possible model, and -1, for a model which completely ignores the evidence.

The first set of runs was made over 109 distributions. Each distribution was generated by a random sample from the eight-dimensional simplex of possible distributions, according to a uniform second-order distribution. For each of these distributions, the $PWR$, $LINR$, $WRST$, $INDP$, and $PRSP$ models were optimally fit and their corresponding $\eta$ values were defined.

## 5. Experimental Results

Using the results of the 109 optimizations, the mean, $\mu$, and standard deviation, $\sigma$, of $\eta$ were calculated for each UIS, and are displayed in table 3.1

|   | INDP | PRSP | PWR |
|---|---|---|---|
| $\mu$ | .8559 | .0711 | .2529 |
| $\sigma$ | .1180 | .1110 | 0.0590 |
| $\mu/\sigma$ | 7.25 | 0.64 | −4.29 |

Figure 3.1 : $\eta$-Performance on uniformly generated distributions

Surprisingly, of the six models tested, the $INDP$ model performed best. Its mean was more than seven standard deviations above the performance of the linear model, indicating a very statistically significant improvement. The $PRSP$ model had a positive mean, but with a comparatively large variance. It was less than one standard deviation above, indicating that it was not significantly more accurate than ordinary linear regression.

---

[4] Deflected gradients is well-known for both quadratic convergence and robustness on difficult problems such as helical ridges. As the usual precaution against local minima, each optimization was done several times from different starting points. To guard against accumulated round-off error, the estimated inverse Hessian matrix was re-initialized to the identity matrix after every $N + 1$ iterations, where $N$ is the number of free parameters. These and more precautions and optimizations are discussed in [1].

An important feature of figure 3.1 is that optimization does not "even things out"; the $PWR$ model did worse than ordinary linear regression. Thus, even after optimization, some seemingly plausible models can not be fixed; their structural form is simply not flexible enough. In that they determine "flexibility", prior assumptions obviously have a strong effect on how well systems perform after optimization, but the effect can not be measured except through extensive testing. Conversely, some models benefit much more from optimization than do others, as is clear from comparison of $INDP$ and $PRSP$, but again testing is necessary to discover this.

One should note that the $PRSP$ model has several advantages over $INDP$, which would cause one to expect better performance from $PRSP$:

- $PRSP$ has seven free parameters, as opposed to the four which are available to $INDP$. In general a model with more degrees of freedom should be capable of finer tuning than a model with few degrees of freedom.
- $INDP$ is structured so that all possible combinations of its parameters form valid probability distributions. This restriction does not apply to $PRSP$, whose seven interdependent parameters are optimized without regard for self-consistency. Of course, if the optimal solution has self-consistent parameters, the optimization routine would find it. Hence, the search for $INDP$ was confined to a more restricted set of parameters, as well as fewer.

The objection immediately arises that the above analysis is unfair to $PRSP$, because $PRSP$ was derived under a certain specific set of assumptions about the prior, which are generally violated when we generate distributions at random. There are at least two responses to this. First, $INDP$ is operating under exactly the same handicap, yet seems to do well. In fact, the assumptions underlying $INDP$ are much more restrictive than those underlying $PRSP$, so we should expect $PRSP$ to perform much better than $INDP$. Second, we can repeat the above analysis, except generating and using only distributions which *exactly* satisfy the assumptions of $PRSP$. Again, this should place $INDP$ at a great disadvantage.

It is important to notice the two sources of error when the analysis is repeated to compare $INDP$ and $PRSP$, using prior distributions designed to meet $PRSP$'s prior assumptions. All the error in $PRSP$ will come from its approximate updating formula, and none at all from its prior assumptions (because they are exactly met). All the error in $INDP$ will come from the prior assumptions, and none at all from its updating formula (because it is exact, given the prior assumptions). When the above analysis is repeated under the $PRSP$ assumptions, we get the table in figure 3.2.

|   | $INDP$ | $PRSP$ | $PWR$ |
|---|---|---|---|
| $\mu$ | .8951 | .5166 | −.3561 |
| $\sigma$ | .0843 | .3488 | .0773 |
| $\mu/\sigma$ | 10.618 | 1.481 | −4.607 |

Figure 3.2 : $\eta$-Performance on conditionally independent distributions

As one can see, $INDP$ *still* outperforms $PRSP$ even when the data has been deliberately biased so as to favor $PRSP$ as much as possible. The statistical significance of the $INDP$'s improvement has increased to almost eleven standard deviations. The mean for $PRSP$ has shifted to a statistically significant improvement over the linear model, but not enormously so. Thus, the assumptions of $PRSP$ have been perfectly satisfied, while those of

$INDP$ have been strongly violated, and yet $INDP$ outperforms $PRSP$ by a wide margin.[1]

It seems that, *after optimization*, $INDP$ is a better model under $PRSP$'s assumptions than is $PRSP$ itself. The $INDP$ model has been roundly criticized for its strict and unrealistic assumptions, but apparently it responds quite well to optimization. Hence, the title of this paper: satisfaction of implicit assumptions about the prior is a poor predictor of final, optimized performance.

How can this possibly be true? The explanation lies in the fact that there are really two possible sources of error here: prior assumptions, and updating. As we mentioned earlier, the $INDP$ model gives exactly the correct odds-ratio update, under the assumption of independence. The $PRSP$ model gives an approximate update, under the assumption of conditional independence. It has been implicitly assumed by many prior researchers that the error induced by approximate assumptions will largely dominate the error induced by approximate updating. These results prove that, *after optimization*, the error due to approximate updating under exact assumptions can outweigh the error due to exact updating under approximate assumptions. Moreover, the combination of complicated updating formulae with optimization makes the results not only analytically untractable but also hard to predict intuitively, as exemplified by the results for the $INDP$ model.

## 6. Limitations of this Work

While these results are interesting, it is important to note limitations of the work. First, we have only examined parts of a few systems, and there are obviously many more systems to be analyzed. While the problems uncovered here may or may not be found elsewhere, it is clear that the influences on performance can be subtle and hence that newer systems will need testing of their final performance.

The numbers we presented represent *average accuracy* for a rule set. While the $LINR$ and $PRSP$ models may have the same average accuracy for a certain rule set, this tells us nothing about whether they are producing similar answers. They might have their errors concentrated in different regions of the input-space, and, if true, this might serve as clues for designing a new, better system. For example, suppose that the $PRSP$ model turned out to be quite accurate when some of the probabilities were near 0 or 1, while the $LINR$ model was accurate in the middle range. We could then define a new model which was a weighted average of $PRSP$ and $LINR$, with $PRSP$ dominating near the extremes and $LINR$ dominating in the middle. Of course, one would have to perform a new optimization on the $PRSP$ parameters, $LINR$ parameters, and weighting function, but the resultant model would clearly be more accurate than either model alone.

Related to the previous point, it would be interesting to see not only average accuracy over multiple problems, but to look at classes of problems where performance was quite good or quite weak. What problem features separate the two classes? Are they the same features for several different UIS's?

While accuracy is a fundamental criterion on which to judge a UIS, it is not the only one, and we have not investigated how these criteria may be traded against each other. We have not addressed the issue of how much inaccuracy should be traded for speed of execution and ease of elicitation, as it will vary widely from case to case. However, it seems clear that, among the models which are roughly as accurate as the linear model, by far the simplest, cheapest, and most modular is the linear model itself. It is also clear that there are definite limits to how far the trade-offs can go. For example, it would seem quite difficult to justify extemely inaccurate models (e.g. less accurate than the linear model) as reasonable approximations which are easy to elicite, and fast to use, simply because they are not good approximations.

---

[1] It is worth noting that the EMYCIN system, under Heckerman's interpretation of certainty factors, implicitly makes the same conditional independence assumptions as does PROSPECTOR [3], and also gives the correct answer t when $e_{1i}$ and $e_{2i}$ are both either 0 or 1. However, we have just demonstrated that "getting the assumptions right" does not assure accuracy. Hence, we do not know how accurate EMYCIN will be under Heckerman's interpretation, but there is no a *priori* reason to suspect that EMYCIN's ad *hoc* interpolation function will be markedly more accurate than PROSPECTOR's interpolation function.

We have not examined large rule sets because our analysis techniques are (so far) very expensive. This is strongly related to the point that our techniques are not ready for real-time use, and are (for now) restricted to the role of off-line standards. We have been able to demonstrate what happens with individual rules, and the real issue thus relates to the propagation of errors. There is some question as to whether errors will cancel or compound when many rules are put together. However, it is again clear that this can have a dramatic effect on performance, and needs to be experimentally analyzed on optimized systems.

Lastly, all our results are numerical not algebraic. While we can get an idea of the general sensitivities from such a numerical analysis, it would be more satisfying to have a closed formula for the error so that we could find its derivative with respect to various parameters.

## 7. Conclusions

We have attempted to show that the current debate over which uncertain inference system is best, or which is a better approximation to complete probability theory, has been overly concerned with the role of "implicit assumptions". This concern makes perfect sense *if* implicit assumptions about the prior are good predictors of both a system's final accuracy and its robustness in the real world. We have proven that this is not generally true, and that satisfaction of assumptions is a poor predictor of final performance. Moreover, we have demonstrated that some popular, well-known systems are no more accurate on the average than a simple linear regression. Even under very favorable conditions, they may not be better than the simple marginal independence assumption. It has *recently* been demonstrated that, if inference is viewed as a game against nature in which one tries to minimized mean squared error, and one can estimate the first and second moments from data, then the linear model is in fact the minimax strategy to take [5]. This result may help explain why the linear model turned out to be so robust. While we have concentrated this analysis on the old system from PROSPECTOR, it is clear that newer systems must have their accuracy and robustness defended by explicit analysis of that performance, not merely by analysis of implicit assumptions.